# Explainable Sentiment Analysis with DeepSeek-R1: Performance, Efficiency, and Few-Shot Learning

Donghao HUANG[1,2], Zhaoxia WANG[1]

[1] School of Computing and Information Systems, Singapore Management University, Singapore

[2] Research and Development, Mastercard, Arlington, VA, USA
dh.huang.2023@smu.edu.sg, zxwang@smu.edu.sg ,

*Abstract—*

*Large language models (LLMs) have transformed sentiment analysis, yet balancing accuracy, efficiency, and explainability remains a critical challenge. This study presents the first comprehensive evaluation of DeepSeek-R1—an open-source reasoning model—against OpenAI's GPT-4o and GPT-4o-mini. We test the full 671B model and its distilled variants, systematically documenting few-shot learning curves. Our experiments show DeepSeek-R1 achieves a 91.39% F1 score on 5-class sentiment and 99.31% accuracy on binary tasks with just 5 shots, an eightfold improvement in few-shot efficiency over GPT-4o. Architecture-specific distillation effects emerge, where a 32B Qwen2.5-based model outperforms the 70B Llama-based variant by 6.69 percentage points. While its reasoning process reduces throughput, DeepSeek-R1 offers superior explainability via transparent, step-by-step traces, establishing it as a powerful, interpretable open-source alternative.*

## Introduction

Sentiment analysis is a fundamental task in natural language processing (NLP) with broad applications in e-commerce, finance, and social media monitoring [1], [2]. Recent advancements in large language models (LLMs), such as OpenAI's GPT-4, have significantly improved sentiment classification accuracy [3]. However, when viewed as a text classification task, the interplay among reasoning capabilities, explainability, and computational efficiency remains underexplored [4].

Traditional sentiment analysis models primarily focus on classifying text into broad sentiment categories—positive, neutral, or negative—without providing insights into the reasoning behind their predictions [5]. With the growing emphasis on explainability in artificial intelligence, there is an increasing demand for models that not only classify sentiment accurately but also justify their decisions in a transparent manner. DeepSeek-R1 [6], an open-source family of reasoning models trained via reinforcement learning, introduces a novel approach by generating step-by-step reasoning processes, distinguishing it from conventional LLMs that prioritize direct responses.

In this study, we conduct a systematic evaluation of DeepSeek-R1, GPT-4o, and GPT-4o-mini on multiple affective computing tasks: Amazon Reviews (5-class sentiment classification), IMDB Movie Reviews (binary classification), and GoEmotions (27-class emotion recognition). We employ few-shot prompting techniques (ranging from 0-shot to 50-shot configurations) to assess the models' in-context learning effectiveness and analyze how each model balances accuracy, reasoning depth, and computational efficiency.

To our knowledge, this is the first in-depth evaluation of DeepSeek-R1's explainable sentiment analysis against proprietary models. Our key contributions are:

- We systematically evaluate DeepSeek-R1 (full 671B model and distilled variants: 8B, 14B, 32B, 70B)

against proprietary models (GPT-4o, GPT-4o-mini) across three sentiment benchmarks: Amazon Reviews (5-class), IMDB Reviews (binary), and GoEmotions (27-class).

- We demonstrate that DeepSeek-R1 delivers transparent, step-by-step reasoning traces while maintaining high accuracy (91.39% F1 on 5-class sentiment), enabling the first direct comparison of explainability between open and proprietary LLMs.
- Through a 0–50-shot evaluation across all tasks, we uncover that shot optima varies substantially across DeepSeek-R1 models, driven by task complexity, base architecture, and model learning capacity.
- We show that distillation effectiveness depends on base architecture: the 32B Qwen2.5-based variant outperforms the 70B Llama-based model by 6.69 percentage point, challenging parameter-centric assumptions.
- Our curated datasets, codebase, and complete evaluation results are publicly available on GitHub[1] to support future research in explainable sentiment analysis and reasoning-capable LLMs.

## Related Work

Large Language Models (LLMs) have redefined sentiment analysis through zero- and few-shot learning. GPT-based models, such as GPT-3 [7] and GPT-4 [8], have shown strong performance in zero-shot and few-shot classification settings. Meanwhile, FLAN-T5 [9] has demonstrated the benefits of instruction tuning for more efficient sentiment analysis.

**Reasoning in LLMs for Sentiment Analysis** Two major paradigms have emerged for enabling reasoning in LLMs for sentiment analysis: *prompted reasoning* and *native reasoning*.

Prompted Reasoning: Chain-of-thought prompting has proven effective for guiding models like GPT-4 through multi-step reasoning tasks [10], often enhanced by techniques such as self-consistency decoding [11]. In sentiment analysis, frameworks like THOR [12] apply this method to decompose sentiment into intermediate reasoning steps, yielding strong performance on implicit sentiment tasks. However, these approaches depend on carefully crafted prompts to elicit reasoning from models that are not inherently designed for it.

Native Reasoning: DeepSeek-R1 [6] takes a fundamentally different approach—explicitly trained via reinforcement learning to generate step-by-step reasoning as part of its output. Unlike closed-source models such as OpenAI's o3 and o4-mini, which provide only summarized reasoning, DeepSeek-R1 produces complete reasoning traces during inference. This enables built-in interpretability and transparent decision-making, which are especially important for tasks like sentiment analysis where intermediate reasoning supports output validation.

**Interpretability in Sentiment Analysis** For interpretability in sentiment analysis, the previous work by Luo et al. [4] discusses the concept of extracting text rationales for sentiment predictions to make them more explainable. Other Recent work by Mabokela et al. [13] integrates pre-trained language models with XAI techniques (e.g., attention visualization, LIME, SHAP) to generate post-hoc explanations for transformer-based sentiment classifiers. These methods typically provide *post-hoc explanations* for model decisions.

Our work bridges these areas by evaluating a reasoning-enabled LLM with *built-in interpretability* on sentiment analysis, distinguishing itself by providing the first comprehensive comparison between DeepSeek-R1's native reasoning capabilities and GPT-4o's prompted reasoning performance. This comparison illuminates the trade-offs between inherent transparency and external explanation methods in sentiment classification tasks.

**Few-Shot Learning in Sentiment Analysis** Few-shot prompting has emerged as a powerful technique to reduce reliance on labeled datasets [14]. Studies such as BdSentiLLM [15] has explored large-scale sentiment classification using minimal examples. However, existing studies have not systematically stretched prompting beyond 20-shot examples, nor have they compared how different reasoning paradigms (native vs. prompted) affect few-shot learning efficiency—gaps this study seeks to address.

## Methodology

### Datasets

We evaluate on three datasets:

- **Amazon Reviews** [16]: A fine-grained sentiment analysis dataset containing 1–5 star ratings mapped to five sentiment classes: "`Strongly Negative`", "`Negative`", "`Neutral`", "`Positive`", and "`Strongly Positive`".
- **IMDB Movie Reviews** [17]: A binary sentiment dataset with labels "`Negative`" and "`Positive`".
- **GoEmotions** [18], [19]: A fine-grained emotion dataset with 27 categories for nuanced sentiment evaluation.

[1] https://github.com/inflaton/Explainable-Sentiment-Analysis-with-DeepSeek-R1

Using the original Amazon Reviews and IMDB Movie Reviews datasets, we constructed a fine-grained dataset via human annotation, consisting of 1,838 Amazon and 1,447 IMDB reviews, respectively. Each dataset was split into 70% training and 30% testing sets using a fixed random seed (42) for reproducibility.

For GoEmotions, we adopt the original splits, exclude entries with multiple labels or `neutral`, and ensure a single emotion label per instance. We then subsample 1,400 training and 600 test examples using the same seed for consistency.

## Model Integration

We compare the following models:

- **DeepSeek-R1 Full Model** (671B parameters)
- **DeepSeek-R1 Variant Models** (8B, 14B, 32B, 70B)
- **OpenAI's GPT-4o and GPT-4o-mini**

The DeepSeek-R1 full model (hereafter referred to as `DeepSeek-R1`) and OpenAI's models were evaluated using their respective official APIs. The distilled DeepSeek-R1 variant models were executed locally using Ollama[2] on a high-performance PC equipped with an NVIDIA RTX A6000 Ada Generation GPU, running Ubuntu 22.04 via the Windows Subsystem for Linux (WSL), is equipped with 48 GB of VRAM.

Throughout this paper, the distilled models are referred to as `deepseek-r1:8b`, `deepseek-r1:14b`, `deepseek-r1:32b`, and `deepseek-r1:70b`, following the official naming conventions provided on the Ollama website[3]. For model inference, we used Ollama's and DeepSeek's OpenAI-compatible APIs for the DeepSeek-distilled and full models, respectively, as well as OpenAI's official API for `GPT-4o` and `GPT-4o-mini`. We adopted OpenAI's default inference settings (e.g., `temperature` = 1.0, `top_p` = 1.0, etc.) to ensure standardized and reproducible results.

## Few-shot Prompting

To improve sentiment classification performance, we employed few-shot prompting techniques. To enable explainable sentiment classification, we carefully designed our system prompt to instruct LLMs to provide an explanation alongside the sentiment classification.

**System Prompt Template** Our system prompt template is structured as follows:

```
You are an advanced sentiment analysis assistant.
↪   Your task is to analyze text and provide a
↪   sentiment rating along with a brief
↪   explanation.

The sentiment rating should be based on a
↪   {scale}-point scale: {sentiments}.

Always respond with a JSON object containing the
↪   sentiment and the explanation.

{exemplars}
```

[2]https://ollama.com/

[3]https://ollama.com/library/deepseek-r1

The placeholders in this template are dynamically populated with dataset-specific values:

- `{scale}`: Set to 5 for the Amazon Review test set, 2 for the IMDB Movie Reviews test set, and 27 for the GoEmotions test set.
- `{sentiments}`: Defined as appropriate sentiment/emotion categories for each dataset.
- `{exemplars}`: Populated differently based on the task configuration (detailed below).

**Exemplar Configuration** For zero-shot tasks, the `{exemplars}` field remains empty. For few-shot tasks, exemplars are randomly selected from the training set with a fixed random seed (42), stratified to ensure representation of all sentiment/emotion classes. The same exemplar set is used for all model evaluations at a given shot count to ensure fair comparison. The exemplars are formatted as structured input-output pairs:

```
Example Inputs and Outputs:

- Input: The only thing good about this movie is
↪   the artwork on the promotion poster by H. R.
↪   Geiger. Anti-nuke protesters who...
- Output:
↪   {"sentiment":"Negative","explanation":""}

- Input: This product exceeded my expectations in
↪   every way possible.
- Output:
↪   {"sentiment":"Positive","explanation":""}

... (additional examples follow the same format)
```

**User Prompt Template** The user prompt template that accompanies each classification request is:

```
- Input: {input}
- Output:
```

where `{input}` is replaced with the text requiring sentiment classification.

## Evaluation Metrics

Model performance was assessed using predictive accuracy on held-out test data, measured by both the weighted F1 score and accuracy through comparisons of LLM outputs with ground truth labels.

Inference speed was evaluated using the following metrics:

1) **Mean Evaluation Time**: The average time (in seconds) required for the LLM to complete a classification task, abbreviated as **Time (s)**. It is

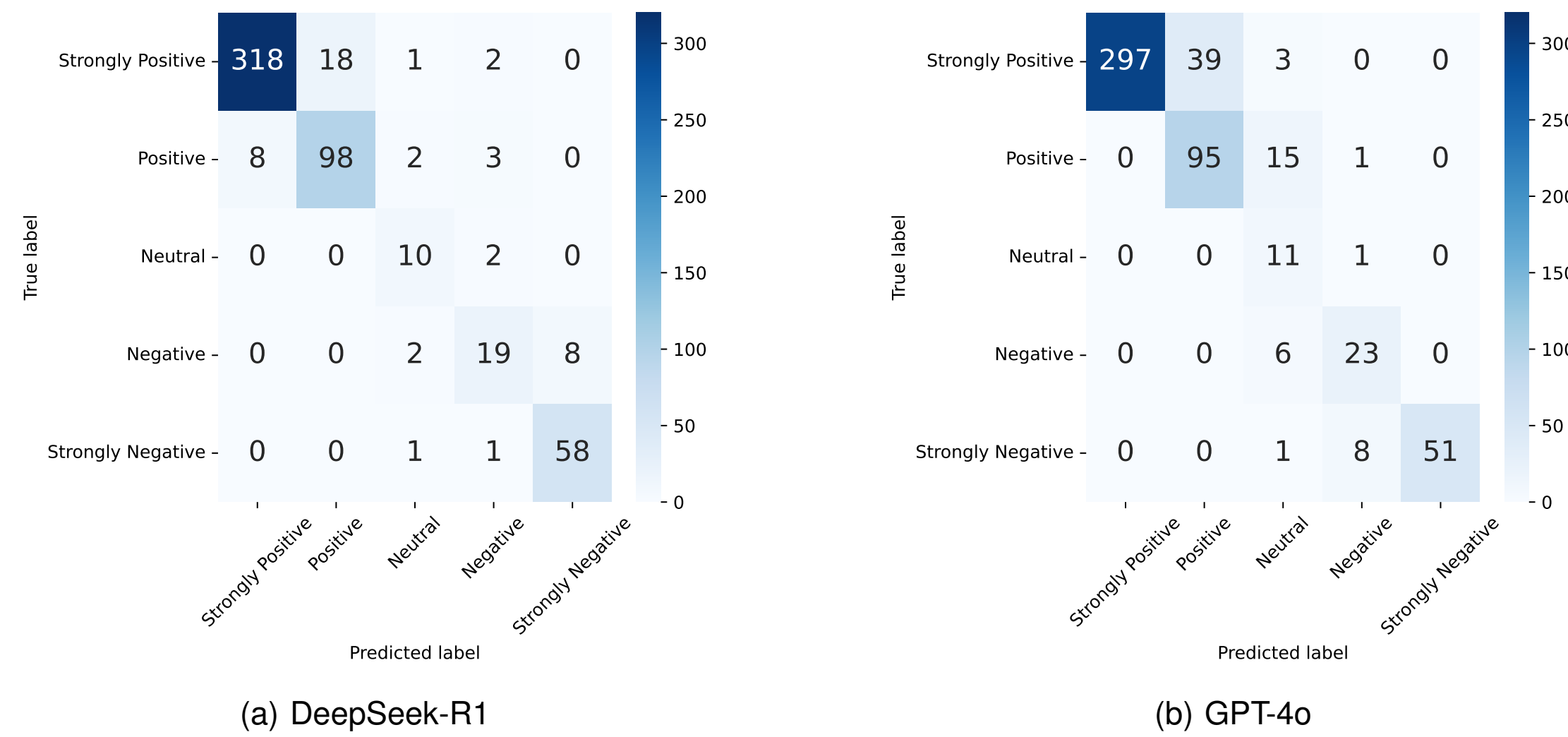

(a) DeepSeek-R1 (b) GPT-4o

FIGURE 1: Fine-Grained Sentiment Analysis: DeepSeek-R1 vs. GPT-4o. Confusion matrices for 5-level sentiment classification on the Amazon Reviews dataset. (a) DeepSeek-R1 exhibits bidirectional confusion between adjacent classes. (b) GPT-4o shows unidirectional confusion, reflecting a conservative treatment of "Strongly" modifiers.

computed as:

$$\text{Mean Evaluation Time} = \frac{\text{Total Evaluation Time (s)}}{\text{Size of Test Dataset}}$$

2) **Throughput**: The number of tokens processed per second, abbreviated as **T-put (t/s)**. It is calculated as:

$$\text{Throughput} = \frac{\text{Total Input \& Output Tokens (t)}}{\text{Total Evaluation Time (s)}}$$

For the **DeepSeek-R1** models (including both full and distilled variants), both the *reasoning content* and the final JSON output are included in the output token count. The *reasoning content* refers to the complete, step-by-step trace of the model's thought process generated during inference, prior to producing the final output. This includes all intermediate thoughts, evaluations, reconsiderations, and decision-making steps. For example, in Table 4, DeepSeek-R1 generates approximately 730 tokens of reasoning (the text under `"Reasoning:"`), followed by a ~90-token JSON output, resulting in a total of around 820 tokens used for throughput calculation. The detailed results will be discussed in the next section, Results and Discussion.

## Results and Discussion

### Comparative Analysis of Model Performance

***Five-Level Sentiment Classification*** Table 1 presents comprehensive results detailing the impact of different shot counts on model performance.

In the five-level sentiment classification task (see Table 1), the full DeepSeek-R1 model achieved the highest performance, reaching 91.29% accuracy and a 91.39% F1 score with 30-shot learning. This significantly outperformed GPT-4o, which achieved 86.57% accuracy and 87.99% F1 at 40-shot, and GPT-4o-mini, which attained 81.31% accuracy and an 83.02% F1 score at 50-shot.

As shown in Table 1, DeepSeek-R1 consistently outperformed all other models from 0-shot to 30-shot, where it reached its peak performance. Notably, it exhibited strong baseline performance in the 0-shot setting (76.77% accuracy, 78.31% F1) and demonstrated superior few-shot learning efficiency, peaking at 30-shot—ten shots earlier than GPT-4o.

Despite its strong performance, DeepSeek-R1 exhibited considerably lower efficiency compared to the OpenAI models, with a throughput of 334 tokens per second (t/s) at its peak (40-shot). In contrast, GPT-4o-mini achieved a throughput of 2,124 t/s, while GPT-4o reached 1,220 t/s. This disparity can primarily be attributed to the design of DeepSeek-R1 as a reasoning model, which leverages extensive test-time compute to enhance its reasoning capabilities.

***Fine-Grained Sentiment Analysis: DeepSeek-R1 vs. GPT-4o*** We compare DeepSeek-R1 and GPT-4o on fine-grained sentiment distinctions—e.g., "Strongly Positive" vs. "Positive"—using Amazon Reviews confusion matrices. Figure 1 reveals their contrasting adjacent-class confusion patterns:

| Model | Metric | 0-shot | 5-shot | 10-shot | 20-shot | 30-shot | 40-shot | 50-shot |
|---|---|---|---|---|---|---|---|---|
| deepseek-r1:8b | Acc (%) | 54.45 | 51.00 | 60.44 | 63.16 | 64.07 | **66.79** | 62.07 |
| | F1 (%) | 58.41 | 56.29 | 65.42 | 67.18 | 68.17 | **69.84** | 66.26 |
| | Time (s) | 2.56 | **2.48** | 2.64 | 2.64 | 2.83 | 2.97 | 3.04 |
| | T-put (t/s) | 82 | 319 | 407 | 643 | 795 | 966 | **1174** |
| deepseek-r1:14b | Acc (%) | 58.44 | 57.17 | 64.61 | 69.33 | **76.95** | 72.41 | 75.14 |
| | F1 (%) | 62.20 | 61.57 | 68.12 | 72.10 | **79.06** | 75.28 | 77.24 |
| | Time (s) | 6.14 | **4.55** | 4.58 | 4.61 | 4.76 | 4.94 | 4.90 |
| | T-put (t/s) | 31 | 172 | 234 | 368 | 473 | 582 | **733** |
| deepseek-r1:32b | Acc (%) | 64.97 | 69.51 | 73.14 | 73.32 | 75.50 | 77.68 | **79.31** |
| | F1 (%) | 69.03 | 72.63 | 75.90 | 75.75 | 77.68 | 79.27 | **80.89** |
| | Time (s) | 9.66 | 8.66 | **8.62** | 8.97 | 8.97 | 8.96 | 9.45 |
| | T-put (t/s) | 20 | 90 | 124 | 189 | 251 | 321 | **380** |
| deepseek-r1:70b | Acc (%) | 60.25 | 62.79 | 67.15 | 68.42 | **71.14** | 67.70 | 69.51 |
| | F1 (%) | 64.34 | 66.75 | 70.50 | 71.70 | **74.23** | 70.63 | 72.59 |
| | Time (s) | 21.96 | 17.73 | 17.21 | **16.86** | 17.59 | 17.25 | 17.85 |
| | T-put (t/s) | 10 | 45 | 62 | 100 | 128 | 166 | **200** |
| DeepSeek-R1 | Acc (%) | 76.77 | 81.13 | 79.67 | 84.39 | **91.29** | 86.57 | 82.58 |
| | F1 (%) | 78.31 | 82.33 | 81.26 | 84.70 | **91.39** | 86.87 | 83.35 |
| | Time (s) | 7.83 | **7.53** | 7.95 | 10.51 | 10.10 | 8.59 | 26.77 |
| | T-put (t/s) | 27 | 105 | 135 | 161 | 223 | **334** | 134 |
| GPT-4o-mini | Acc (%) | 71.87 | 71.14 | 76.95 | 80.76 | 80.58 | 79.67 | **81.31** |
| | F1 (%) | 74.75 | 74.15 | 79.25 | 82.50 | 82.37 | 81.57 | **83.02** |
| | Time (s) | 1.72 | **1.35** | 1.38 | 1.62 | 1.69 | 1.72 | 1.68 |
| | T-put (t/s) | 121 | 587 | 776 | 1046 | 1329 | 1666 | **2124** |
| GPT-4o | Acc (%) | 71.32 | 68.78 | 70.24 | 78.04 | 79.49 | **86.57** | 81.67 |
| | F1 (%) | 74.32 | 72.39 | 73.73 | 80.43 | 81.76 | **87.99** | 83.11 |
| | Time (s) | 1.70 | **1.60** | 1.70 | 2.13 | 2.57 | 2.35 | 4.85 |
| | T-put (t/s) | 123 | 492 | 632 | 795 | 876 | **1220** | 737 |

TABLE 1: Performance on Amazon Reviews Dataset (5-level classification) across different shot counts

**DeepSeek-R1** shows balanced bidirectional confusion:

- Strongly Positive → Positive: 18/339 (5.3%)
- Positive → Strongly Positive: 8/111 (7.2%)
- Strongly Negative → Negative: 1/60 (1.7%)
- Negative → Strongly Negative: 8/29 (27.6%)

**GPT-4o** adopts a strictly conservative strategy:

- Strongly Positive → Positive: 39/339 (11.5%)
- Positive → Strongly Positive: 0/111 (0%)
- Strongly Negative → Negative: 8/60 (13.3%)
- Negative → Strongly Negative: 0/29 (0%)

DeepSeek-R1 demonstrates greater flexibility, with 2.2× fewer misclassifications on "Strongly Positive" (5.3% vs. 11.5%). Unlike GPT-4o, it also correctly upgrades sentiment when appropriate. For positive sentiment, DeepSeek-R1 shows symmetric confusion (5.3% ↓, 7.2% ↑), while GPT-4o never upgrades (11.5% ↓, 0% ↑). For negative sentiment, DeepSeek-R1 struggles more with intensity upgrades (27.6%), whereas GPT-4o remains conservative (13.3% ↓, 0% ↑).

This analysis suggests that DeepSeek-R1's reasoning-oriented architecture enables finer-grained sentiment distinctions, particularly for intensity modifiers, while GPT-4o favors conservative and consistent assignment.

***DeepSeek Variant Performance Analysis*** Our evaluation reveals substantial performance differences among DeepSeek-R1 variants that extend beyond parameter scaling, highlighting the interplay between architecture, size, and task-specific outcomes.

The Qwen2.5-based 14B and 32B models consistently outperform their Llama-based counterparts: the 32B variant achieves 80.45% F1, exceeding the 70B model's 73.76% by 6.69 percentage points, demonstrating that architectural design plays a central role in distillation effectiveness. Performance does not scale monotonically with model size; despite having over twice the parameters, the 70B model underperforms the 32B variant, indicating diminishing returns from the Llama3.3-70B base.

Regarding throughput–accuracy trade-offs, the 8B variant processes 1,403 tokens/s with a 69.61% F1, making it ideal for high-volume tasks; the 14B model balances efficiency and accuracy at 881 tokens/s and 78.50% F1; the 32B variant offers the best trade-off at 457 tokens/s and 80.45% F1; and the 70B model, despite its higher cost (239 tokens/s), yields only 73.76% F1, reflecting limited performance gains.

Deployment implications vary by use case: for high-throughput scenarios, the 8B variant offers practical accuracy with 3.1× the throughput of the 32B model;

for balanced deployments, the 32B variant achieves 88.0% of full-model accuracy with significantly greater efficiency; and for accuracy-critical tasks, the full DeepSeek-R1 justifies its computational overhead by delivering maximum performance. Overall, model selection should be guided by task complexity, accuracy requirements, and resource constraints—not parameter count alone.

***Binary Sentiment Classification*** To establish baseline performance on fundamental sentiment analysis, we evaluated all models on the IMDB Movie Reviews dataset for binary sentiment classification. This task serves as a foundational benchmark for understanding model capabilities on sentiment-related tasks before progressing to more complex emotion recognition.

Table 2 presents the performance of DeepSeek-R1 variants, GPT-4o-mini and GPT-4o on binary sentiment classification. The results demonstrate that this task represents a relatively solved problem for modern large language models, with most variants achieving excellent performance.

| Model | Acc (%) | F1 (%) | Shots | Time (s) | T-put (t/s) |
|---|---|---|---|---|---|
| deepseek-r1:8b | 95.17 | 95.71 | 10 | 3.21 | 1103 |
| deepseek-r1:14b | 97.47 | 97.47 | 20 | 5.76 | 1133 |
| deepseek-r1:32b | 96.55 | 97.11 | 40 | 12.83 | 918 |
| deepseek-r1:70b | 97.93 | 98.04 | 20 | 89.66 | 73 |
| DeepSeek-R1 | **99.31** | **99.31** | 5 | 20.14 | 143 |
| GPT-4o-mini | 99.08 | 99.08 | 30 | 2.17 | 5574 |
| GPT-4o | **99.31** | **99.31** | 40 | 2.64 | 5350 |

TABLE 2: Performance on IMDB Movie Reviews dataset (binary sentiment classification) at optimal few-shot settings

Both DeepSeek-R1 and GPT-4o achieve exceptional performance at 99.31% accuracy and F1 score, demonstrating near-perfect sentiment classification capabilities. Remarkably, DeepSeek-R1 reaches this performance with only 5 shots, while GPT-4o requires 40 shots to achieve the same level. This 8× improvement in few-shot efficiency highlights DeepSeek-R1's superior reasoning capabilities for sentiment classification tasks.

The performance scaling across DeepSeek-R1 model sizes shows consistent improvement, with accuracy increasing from 95.17% for the 8B variant to 99.31% for the full model. Even the smaller variants demonstrate strong performance, with the 14B model achieving 97.47% accuracy while maintaining excellent throughput (1133 t/s). The efficiency advantage of GPT-4o is evident in the throughput metrics, with GPT-4o achieving 5350 t/s compared to DeepSeek-R1's 61 t/s, though both models deliver equivalent accuracy on this fundamental sentiment analysis task.

***Fine-Grained Emotion Recognition*** To assess model performance beyond coarse sentiment polarity, we evaluate on the GoEmotions dataset [18], which comprises 27 emotion categories—a substantially more challenging task than binary or 5-class sentiment classification. Table 3 presents results for DeepSeek-R1 variants, GPT-4o-mini and GPT-4o. While all models show performance drops relative to simpler sentiment tasks, DeepSeek-R1 remains competitive with GPT-4o on this fine-grained benchmark.

| Model | Acc (%) | F1 (%) | Shots | Time (s) | T-put (t/s) |
|---|---|---|---|---|---|
| deepseek-r1:8b | 32.17 | 32.24 | 50 | 4.61 | 519 |
| deepseek-r1:14b | 36.83 | 38.69 | 40 | 4.45 | 414 |
| deepseek-r1:32b | 39.83 | 41.14 | 50 | 8.43 | 254 |
| deepseek-r1:70b | 44.00 | 44.57 | 10 | 17.25 | 52 |
| DeepSeek-R1 | **45.33** | **46.02** | 10 | 28.79 | 48 |
| GPT-4o-mini | 39.83 | 40.52 | 5 | 1.34 | 340 |
| GPT-4o | 45.17 | 45.99 | 10 | 1.43 | 424 |

TABLE 3: Performance on GoEmotions dataset (27-class emotion recognition) at optimal few-shot settings

The full DeepSeek-R1 achieves the highest scores (45.33% accuracy, 46.02% F1), slightly surpassing GPT-4o. Both models reach peak performance with 10-shot prompts, indicating strong few-shot generalization in this complex task.

Few-shot trends show variant-specific behavior: 8B and 32B achieve peak performance at 50 shots, while 70B requires only 10. GPT-4o-mini peaks at 5 shots with lower accuracy (39.83%), underperforming compared to deepseek-r1:32b, indicating limited benefit from additional samples.

Overall, performance on GoEmotions is notably lower than on sentiment tasks, reflecting the added challenge of classifying 27 nuanced emotion classes.

***Impact of Few-Shot Examples*** Our analysis shows that the optimal number of few-shot examples varies substantially across DeepSeek-R1 models, driven by task complexity, base architecture, and model learning capacity. Figure 2 summarizes performance trajectories across shot counts for each model and dataset.

**Task Complexity Effects** Optimal shot counts correlate strongly with task complexity. IMDB (Figure 2a) requires the fewest examples (5–20 shots), while Amazon's 5-level sentiment task (Figure 2b) typically peaks at 30–50 shots. GoEmotions (Figure 2c), the most complex task, shows high variance (10–50 shots), reflecting the challenges of fine-grained emotion recognition.

**Architecture-Specific Learning Patterns** Model behavior varies by base architecture. Qwen2.5-based models (14B and 32B) show consistent patterns: 14B peaks at 20–40 shots, while 32B often needs 40–

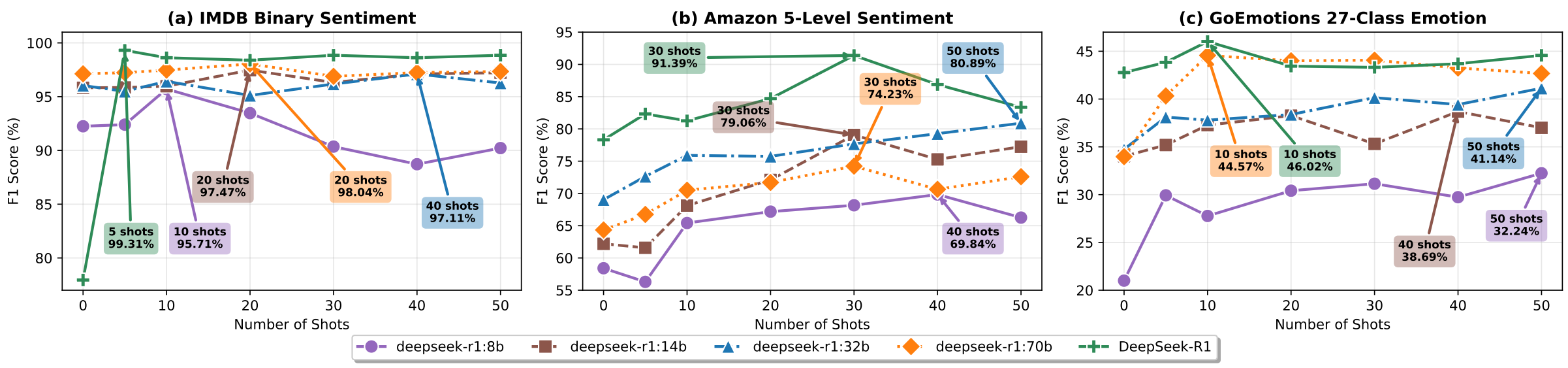

FIGURE 2: Performance of DeepSeek-R1 variants across different few-shot configurations. (a) IMDB Binary Sentiment: rapid convergence for most models. (b) Amazon 5-Level Sentiment: higher shot requirements for fine-grained classification. (c) GoEmotions 27-Class Emotion: large variance in optimal shot counts across models. Peak annotations show the best-performing shot count and corresponding F1 score.

50. Llama3.x-based models (8B and 70B) are more variable. The 8B model scales predictably with task complexity—peaking at 10, 40, and 50 shots across the three tasks—suggesting smaller models rely more heavily on in-context examples. The full DeepSeek-R1 model achieves the highest overall performance with the lowest average shot count (15.0) and largest average F1 gain (12.56%). This reflects the advantage of its full reasoning architecture over distilled variants.

**Model Learning Efficiency Insights** Larger performance gains often require more examples. The 8B model shows the second-highest average improvement (10.62%), needing more support for complex tasks. In contrast, the 14B and 32B models start strong but show diminishing returns (7.74% and 6.63% average improvements, respectively), indicating a ceiling effect tied to model capacity.

These findings suggest that optimal shot counts reflect the interplay between model limitations and learning efficiency: smaller models benefit more from added context, while larger ones approach peak performance with fewer examples—especially in simpler tasks (e.g., binary sentiment classification tasks).

## Explainability in Sentiment Classification

As discussed in the *Few-Shot Prompting* section, we designed our system prompt to guide LLMs in providing explanations alongside sentiment classifications. Additionally, we capture the reasoning content from the DeepSeek-R1 models to further enhance interpretability. To illustrate this, we present three diverse cases in Table 4, Table 5, and Table 6.

These examples highlight key differences between GPT-4o and DeepSeek-R1 in their approaches to sentiment analysis and explainability.

***Resolution of Mixed Experiences*** In the first example (Table 4), both models correctly identify the neutral sentiment but follow distinct reasoning paths. GPT-4o balances product failure against positive customer service, while DeepSeek-R1 traces the temporal progression from a negative experience to its resolution. This case demonstrates both models' ability to handle evolving sentiments and weigh competing factors.

***Emotional vs. Functional Analysis*** The second example (Table 5) highlights differences in processing emotional cues. GPT-4o weighs functional benefits against emotional disappointment over pricing, maintaining a "`Positive`" rating. In contrast, DeepSeek-R1 assigns greater weight to the emotional impact of feeling "duped," leading to a "`Negative`" sentiment despite acknowledging the product's positive attributes. This underscores DeepSeek-R1's heightened sensitivity to emotional language.

***Treatment of Indirect Experiences*** The third example (Table 6) showcases divergent approaches to reported issues. GPT-4o interprets widespread user complaints as indicative of negative sentiment, even without direct personal experience. DeepSeek-R1, however, classifies the review as "`Neutral`", treating it as primarily informational since it focuses on reporting issues rather than expressing direct dissatisfaction.

***Explainability Approaches*** Both models produce explanations, but their stylistic differences stem from their architectural designs rather than prompt interpretation. As shown in Tables 4–6, both GPT-4o and DeepSeek-R1 follow our prompt for “a brief explanation,” yielding concise summaries in their main outputs.

The key stylistic differences between DeepSeek-R1 and and GPT-4o explanations are that DeepSeek-R1 generates an *intrinsic reasoning trace*—not prompted, but inherent to its architecture. This reasoning trace represents a unique form of built-in explainability that emerges naturally from DeepSeek-R1's reinforcement learning training for reasoning. Unlike GPT-4o's single-layer explanations, DeepSeek-R1 provides dual-layer explanation transparency: it combines a brief, prompt-compliant explanation with an extensive step-by-step trace that reveals the full decision-making process, including moments of uncertainty ("Wait, but the output examples show..."), self-correction, and reconsideration.

These traces offer practical benefits for sentiment analysis tasks. As shown Table 4-6, DeepSeek-R1's reasoning unfolds in multiple stages, allowing users to observe how it reconciles conflicting signals before reaching a final decision. The reasoning trace separates reported problems from personal experiences, indicating a nuanced interpretation that often remains hidden in single-layer explanations.

In summary, DeepSeek-R1's detailed reasoning process enhances transparency by explicitly addressing sentiment ambiguities (e.g., "Great storage size" and "excellent" vs. "got duped" as shown in Table 5) that simpler models may overlook. It systematically evaluates conflicting cues, making its decision-making process more interpretable.

## Conclusion

We have presented the first end-to-end evaluation of DeepSeek-R1—its full 671B model and distilled 8B–70B variants—against GPT-4o and GPT-4o-mini on three sentiment benchmarks. Our key findings are:

- DeepSeek-R1 matches or exceeds proprietary LLMs across Amazon (5-class), IMDB (binary), and GoEmotions (27-class) tasks.
- DeepSeek-R1 delivers native, step-by-step reasoning traces with minimal accuracy loss (91.39% F1 on five-class sentiment).
- Distillation quality depends more on base architecture than size—the 32B Qwen2.5 variant outperforms the 70B Llama model by 6.69 pp.
- Distilled variants offer distinct accuracy–throughput profiles, enabling tailored choices from high-volume (8B) to accuracy-critical (full) settings.

Our findings set new benchmarks for reasoning-based sentiment analysis, demonstrating that open-source LLMs can match proprietary systems in both accuracy and interpretability—albeit at higher computational cost.

Future work will focus on improving inference efficiency (e.g., via specialized hardware), extending to aspect-based sentiment analysis for richer explanations, and refining few-shot exemplar selection to maximize performance. Integration of quantitative explainability metrics into the model explainability analysis will also be of interest for future work.

Overall, this study underscores the trade-offs between reasoning depth, efficiency, and accuracy, and positions DeepSeek-R1 as a transparent, cost-effective alternative for real-world sentiment analysis.

| Review Text & Ground Truth | **Text:** I join the long list of customers who have had their SanDisk die on them. Customer service from SanDisk has been helpful. They pay for UPS shipping for the return and promise a 7-10 day turn around once the defective disk has been returned. Still it's a pain to deal with and I lost some photographs. UPDATE: SanDisk sent me a replacement and it seems to be working. If the replacement goes bad, I'll be back to complain:) SECOND UPDATE: All is well with the replacement so I'm giving it three stars.<br>**Sentiment:** Neutral |
|---|---|
| **GPT-4o Analysis** | **Sentiment:** Neutral<br>**Explanation:** The text provides both negative and positive aspects: the original SanDisk product failed, causing inconvenience and loss of photos, which is a negative point. However, the customer service is noted as helpful, offering a replacement that currently works, leading to a neutral sentiment overall with a moderate satisfaction level. |
| **DeepSeek-R1 Analysis** | **Sentiment:** Neutral<br>**Explanation:** The user experienced significant issues with the product failing and data loss, which is strongly negative. However, SanDisk's customer service was responsive, providing a replacement that worked, leading to a resolution. The mixed experience, with both negative and positive aspects, along with the final three-star rating, results in a Neutral sentiment overall.<br>**Reasoning:** Okay, let's tackle this sentiment analysis. The input is a user's experience with a SanDisk product that failed. They mention joining others who had similar issues, which immediately sounds negative. The customer service was helpful, covering shipping and replacing the product, which is a positive point. However, the user still had to go through the hassle of returning it and lost some photos, which is frustrating. The first update mentions the replacement works, but there's a conditional statement about complaining again if it fails, indicating lingering doubt. The second update says all is well, but they only give three stars, which is average.<br>So, the overall sentiment here is mixed. There are both negative experiences (product failure, lost data) and positive aspects (good customer service, replacement working). However, the user's final rating is three stars, which is neutral. But considering the initial strong negative experience and the hassle involved, even with a resolution, the sentiment might lean more towards Neutral. Wait, but the output examples show that even with some positive elements, if the main issue is resolved but there's lingering dissatisfaction, it might be Neutral. Let me check the examples.<br>Looking at the example where the user had issues but then got a replacement and rated three stars, the output was Neutral. Here, the user mentions the replacement working but still only gives three stars. The main issues were resolved but the initial problem was significant. So the overall sentiment here is Neutral. But wait, in the output example where the user had a bad experience but then a refund and replacement, the sentiment was Strongly Negative. Hmm, maybe because the initial problem was severe.<br>Wait, in the input provided, the user's final rating is three stars, which is in the middle. The explanation includes both negative and positive elements. In some cases, mixed experiences are Neutral. Let's compare with the example where the user had a first defective card, then a working replacement, and the output was Neutral. So in this case, the input here is similar. The user had a problem, got it resolved, but still had some hassle. The final outcome is okay, but the experience was mixed. Therefore, the sentiment should be Neutral. But wait, the initial problem was severe (lost photos, hassle) and even though resolved, the user might still have a negative sentiment. However, the three stars and the fact that they're giving an update that's positive but not overly so might balance it. Let me check the example again.<br>In one of the examples, the user had a defective card, got a refund, replaced with Samsung, and the sentiment was Strongly Negative. But in another example where the user had a problem but the replacement worked, the sentiment was Neutral. So the key difference is whether the problem was resolved satisfactorily. In this input, the user's final update is positive (all is well) but the overall experience was a hassle. The three stars indicate a neutral stance. Therefore, the sentiment here should be Neutral. So the output should be "sentiment":"Neutral","explanation":"" |

TABLE 4: Analysis of Sentiment Classification from GPT-4o and DeepSeek-R1 - Neutral

| | |
|---|---|
| **Review Text & Ground Truth** | **Text:** Great storage size. Being the first adopter when it came out. Thought the sale price $119 is excellent. Got duped. It's been the same price since it was first introduced. Great for the action cam. Battery ran out before it ran out of memory.<br>**Sentiment:** Strongly Positive |
| **GPT-4o Analysis** | **Sentiment:** Positive<br>**Explanation:** The text expresses satisfaction with the storage size and its performance with the action cam, even though there is a disappointment about the price not being a sale deal. The positive aspects of the product overshadow the issue with the pricing. |
| **DeepSeek-R1 Analysis** | **Sentiment:** Negative<br>**Explanation:** The user expresses positive aspects like 'Great storage size' and functionality with the action cam, but the strong negative feeling of being 'duped' due to the unchanging price and initial perceived value disappointment shifts the sentiment to Negative."<br>**Reasoning:** Okay, let's analyze this input step by step. The user starts by mentioning "Great storage size," which is a positive point. They also refer to themselves as the first adopter and thought the sale price of $119 was excellent. However, they then say "Got duped," indicating a negative feeling, because the price hasn't changed since introduction. They mention it's great for the action cam and that the battery died before the memory was full, which is a practical positive but maybe a minor point.<br>The initial positive aspects are countered by the feeling of being misled about the price. The negative emotion from feeling duped seems strong, but the user still acknowledges the product's effectiveness. However, the overall sentiment is more on the negative side because the main issue is the price deception. But the user also has some positive remarks. However, the term "duped" is quite strong, leading towards a negative sentiment. Yet, they end with a positive note about the product's performance. But the primary issue here is the perceived deception, which might tip the sentiment to Negative rather than Neutral. The presence of both positive and negative aspects but with a stronger negative emotion due to feeling tricked. |

TABLE 5: Analysis of Sentiment Classification from GPT-4o and DeepSeek-R1 - Strongly Positive

| | |
|---|---|
| **Review Text & Ground Truth** | **Text:** I had wanted to get the most I can out of my hero3, At the time I got my hero3 this was the recommended card, but there were many problems with people having issues with it. Since then Gopro has endorsed other card and from what I understand changed the firmware to prevent problems.... The problems is only with the 64gb version, from what I read the 32gb and lower versions are ok...<br>**Sentiment:** Strongly Negative |
| **GPT-4o Analysis** | **Sentiment:** Negative<br>**Explanation:** The review discusses issues associated with using a specific memory card with the GoPro Hero3 camera. Although the reviewer highlights that the problems are mainly with the 64GB version and not the smaller capacities, the overall tone suggests dissatisfaction and problems with the recommended product. The reference to many people experiencing issues contributes to the negative sentiment. |
| **DeepSeek-R1 Analysis** | **Sentiment:** Neutral<br>**Explanation:** The input mentions known issues with the 64GB version of the card but clarifies that smaller capacities are acceptable. The tone is informational rather than expressing personal dissatisfaction, as the user references firmware fixes and GoPro's updated endorsements without indicating direct negative experiences.<br>**Reasoning:** Okay, let's tackle this sentiment analysis. The input here is a user talking about their experience with a 64GB memory card for a GoPro Hero3. They mention that it was the recommended card at the time but had issues. They note that GoPro has since endorsed other cards and updated firmware to prevent problems, specifically pointing out that the 64GB version had issues while 32GB and lower are okay.<br>First, I need to determine the overall sentiment. The user starts by stating they wanted the best for their camera and initially got the recommended card. However, they then mention problems that others experienced. The key here is that the user isn't directly reporting a personal negative experience but is referring to known issues with the 64GB version. They also mention that GoPro addressed the problem with firmware and now endorses other cards, implying that the 64GB might still be problematic.<br>The user isn't expressing personal dissatisfaction but is providing cautionary information. They highlight that lower capacity cards are okay, which could be a helpful note for others. There's no strong positive or negative emotion here; it's more of an informative statement. The mention of problems is balanced with the resolution (firmware update) and clarification about other sizes. So the sentiment is likely Neutral because there's no direct complaint or endorsement, just factual information about past issues and current status. |

TABLE 6: Analysis of Sentiment Classification from GPT-4o and DeepSeek-R1 - Strongly Negative

**DONGHAO HUANG** is Vice President of R&D at Mastercard, 4250 Fairfax Drive, 22203, Arlington, VA 22203, USA, leading AI, Machine Learning, and Generative AI research, with a focus on large language models for enterprise applications. He is pursuing a Doctor of Engineering at Singapore Management University. Contact him at: donghao.huang@mastercard.com.

**ZHAOXIA WANG** is an associate professor of computer science (practice) in the School of Computing and Information Systems Singapore Management University, 178902, Singapore. Her research interests include Natural Language Processing (NLP), machine learning, sentiment analysis, explainable AI, large language models (LLMs), and other AI-related research and applications. She is an IEEE Senior Member. Contact her at: zxwang@smu.edu.sg.